\pdfoutput=1

\documentclass[11pt]{article}

\usepackage[final]{acl}

\usepackage{times}
\usepackage{latexsym}
\usepackage{multirow}
\usepackage{booktabs}
\usepackage{lingmac}
\usepackage{tcolorbox}
\usepackage{xcolor}
\usepackage{amssymb}
\usepackage{times}
\usepackage{xspace}

\newcommand{\tss}[1]{\textsuperscript{#1}}
\newcommand{\tn}[1]{\textnormal{#1}}

\usepackage{enumitem}  %
\usepackage{xcolor}    %
\usepackage{tcolorbox} %
\usepackage{soul}   
\newcommand{\hlc}[2][yellow]{{%
    \colorlet{foo}{#1}%
    \sethlcolor{foo}\hl{#2}}%
}

\usepackage{todonotes}

\definecolor{ForestGreen}{RGB}{20,150,20}
\usepackage{fontawesome5} %
\newcommand{\reasoning}{\faBrain}   %
\newcommand{\openmodel}{\faUnlock}    %

\usepackage[T1]{fontenc}

\usepackage[utf8]{inputenc}
\usepackage{microtype}
\usepackage{inconsolata}

\usepackage{graphicx}

\title{Zero-Shot Belief: A Hard Problem for LLMs}

\author{
    John Murzaku\tss{\tn{$\blacklozenge\spadesuit$}},
    Owen Rambow\tss{\tn{$\clubsuit\spadesuit$}} \\
    \tss{$\blacklozenge$}Department of Computer Science
    \tss{$\clubsuit$}Department of Linguistics\\
    \tss{$\spadesuit$}Institute for Advanced Computational Science\\
    Stony Brook University\\
    \texttt{jmurzaku@cs.stonybrook.edu}
}

\begin{document}
\maketitle

\begin{abstract}
We present two LLM-based approaches to zero-shot source-and-target belief prediction on FactBank:  a unified system that identifies events, sources, and belief labels in a single pass, and a hybrid approach that uses a fine-tuned DeBERTa tagger for event detection.
We show that multiple open-sourced, closed-source, and reasoning-based LLMs struggle with the task.
Using the hybrid approach, we achieve new state-of-the-art results on FactBank and offer a detailed error analysis. Our approach is then tested on the Italian belief corpus ModaFact. 
\end{abstract}
\section{Introduction}
The term “belief''
(interchangeably referred to as ``event factuality'' in NLP) refers to the extent an event mentioned by the author or by sources in a text is presented as being factual.  While this task has received attention over the years, no zero-shot experiments have been performed.  We show that this task remains a hard task for LLMs.

Our major contributions are as follows:
\newline (1) We present unified and hybrid zero-shot frameworks 
for the source-and-target belief prediction task (i.e., who has what belief towards what). We test our approach on various LLMs.
\newline (2) Our hybrid approach achieves new state-of-the-art results (SOTA) on the FactBank corpus, but the problem is far from solved.
\newline (3) We are the first to evaluate FactBank on Nested belief, revealing that LLMs perform particularly poorly on this task.  We perform an error analysis showcasing where LLMs fail. 
\newline (4) We validate the transferability of our approach by testing on the Italian ModaFact belief corpus. 

This paper is organized as follows: we provide an  overview of the belief detection task in Section~\ref{sec:relwork}. We follow by detailing our methodology in Section~\ref{sec:methods}
and discuss our results and analysis for FactBank and ModaFact in Section~\ref{sec:results}. 

\section{Related Work}
\label{sec:relwork}
\noindent\textbf{Corpora } Many corpora explore the notion of belief on the sentence level. FactBank is one of the first corpora to do this, annotating source-and-target belief: both the belief presented by the author towards an event and the belief towards events by sources mentioned inside of the text \citep{sauri2009factbank}.
Other corpora annotate only the author's belief towards events: these corpora include LU \cite{diab-etal-2009-committed}, UW \cite{lee-etal-2015-event}, LDCCB \citep{prabhakaran-etal-2015-new}, MEANTIME \citep{minard-etal-2016-meantime}, MegaVeridicality \cite{white-etal-2018-lexicosyntactic},  UDS-IH2 \citep{rudinger-etal-2018-neural-models}, CommitmentBank \citep{de2019commitmentbank}, and RP \citep{ross-pavlick-2019-well}. Two recent corpora for event factuality are Maven-Fact \citep{li-etal-2024-maven} which contains a large-scale corpus of event and supporting evidence annotations, 
and ModaFact \citep{rovera-etal-2025-modafact}, which is an Italian author belief corpus that annotates in a similar style and inspiration as FactBank. 

\noindent\textbf{Methods} Previous methods for author belief prediction mainly involve fine-tuning: \citet{rudinger-etal-2018-neural-models} fine-tune multi-task LSTMs; \citet{pouran-ben-veyseh-etal-2019-graph} fine-tune a graph convolutional network with BERT \citep{devlin-etal-2019-bert} representations; \citet{jiang-de-marneffe-2021-thinks, murzaku-etal-2022-examining} fine-tune RoBERTa \citep{liu2019roberta} with span representations; \citet{li-etal-2024-maven} fine-tune RoBERTa and Flan-T5 \citep{chung2024scaling}, and also explore four LLMs predictions using few-shot learning; \citet{rovera-etal-2025-modafact} fine-tune BERT, mT5-XXL \citep{xue-etal-2021-mt5}, Aya23-8B \citep{aryabumi2024aya}, and Minerva-3B \citep{orlando2024minerva}.

There has been much less focus on the complete source-and-target belief task: \citet{murzaku-etal-2023-towards, murzaku-rambow-2024-beleaf} both fine-tune a Flan-T5 model, with the latter optimizing for the structure of belief represented as a tree.

\section{Preliminaries} 
\label{sec:prelims}
Consider this sentence: 
{\em Trurit Inc. said it is phasing out legacy routers.} This sentence reports on two events: a
``said'' event and a ``phasing'' event. 

\noindent\textbf{Author Belief } The definition of author belief (also called event factuality) %
is how committed is the author of the text (the source) to the truth (or factuality) of an event. In this sentence, the author is presenting the ``said'' event as
factual, i.e., they are committed to the ``said''
event having happened. On the other hand, the author is presenting the ``phasing'' event as having an unknown factuality; the author is not directly committing to the truth of the event, rather they are reporting on what ``Trurit Inc.'' said.

\noindent\textbf{Nested } In nested belief, we report the belief towards events according to nested sources inside of a text. 
The task can be split into three steps: (i) identifying the nested or attributed source in the text; (ii) linking the source to the events (i.e., which events does the source commit to); (iii) labelling the belief of the event according to that source. In our example, the source is ``Trurit Inc.''
Once the source is introduced (i), we then link the source to the events in the text (ii): in this case, Trurit Inc. is reportedly committing to the event ``phasing'', and asserting it as true (iii). Since the source is reporting about this event, and directly committing to the event happening, it is therefore true in Trurit Inc.'s perspective as reported by the author, and unknown in the author's perspective.

\section{Methodology}
\label{sec:methods}
We use the test set of the source-and-target (author and nested sources) projection of FactBank %
released by \citet{murzaku-rambow-2024-beleaf}. 
Further dataset details are in Appendix~\ref{sec:appendix-data}.

\subsection{Zero-Shot}
\noindent\textbf{Unified} Our \textbf{Unified} approach provides a single end-to-end zero-shot prompt to the LLM with the input text, a high-level descriptions of the task, the three main steps in the annotation process in detail, special cases guidelines, and the output format. We end the prompt with a summary of the specific steps on how to produce the final answer in a chain-of-thought format (CoT) \citep{wei2022chain}, which has proven to work well for author event factuality \citep{li-etal-2024-maven}. The three steps are: 
(1) Label all events according to the FactBank annotation guidelines, which we provide.
(2) Identify all nested sources in the text. 
(3) Assign factuality labels for each event, according to that source.
We leave all model details, API parameters, and our exact prompts in Appendix~\ref{sec:appendix-unified}.

\noindent\textbf{Hybrid}
For our \textbf{Hybrid} zero-shot approach, we first extract events in a sentence using a DeBERTa \citep{he2021debertav3} based tagger. After extracting the events, we prompt an LLM with the sentence and the list of events. We then follow the exact steps (minus event detection, since we provide events) as our \textbf{Unified} prompt: we instruct the LLM to identify all nested sources, ask the LLM to assign factuality labels for the events, according to the identified sources, and finish with instructions for answering with CoT.  See Appendix~\ref{sec:appendix-hybrid} for further details on our \textbf{Hybrid} experiments and our exact prompts. 

\noindent\textbf{Event Tagger} The FactBank corpus has a complex definition of what exactly is an annotatable event. \citet{murzaku-etal-2023-towards} found that annotating FactBank events is non-trivial, even with a specialized, generative fine-tuned model achieving only 85.4\% F1 on event identification. We therefore choose to fine-tune a DeBERTa model for event token detection, and then pass the events to our \textbf{Hybrid} prompts.

\subsection{Models}
We perform experiments on a variety of LLM types: open LLMs, specifically LLaMA-3.3-70B \cite{llama}, DeepSeek-v3 \citep{liu2024deepseek}, and DeepSeek-r1; closed LLMs, specifically GPT-4o \citep{gpt4o}, newly released reasoning models o1 \citep{o1} and o3-mini \citep{o3-mini}, and Claude 3.5 Sonnet \citep{claude}; and reasoning LLMs, DeepSeek r1 (henceforth \textsc{r1}), o1, and o3-mini.

\begin{table*}[t]
\setlength\tabcolsep{3pt}
\centering
\begin{tabular}{l c c c | c c c | c c | c c c}
\toprule
\multirow{2}{*}{\textbf{Model}} 
  & \multicolumn{3}{c|}{\textbf{Unified}} 
  & \multicolumn{3}{c|}{\textbf{Hybrid}} 
  & \multicolumn{2}{c|}{\textbf{\(\Delta\) Hyb.-SOTA}} 
  & \multicolumn{3}{c}{\textbf{\(\Delta\) Hyb.-Unif.}} \\
\cmidrule(lr){2-4}\cmidrule(lr){5-7}\cmidrule(lr){8-9}\cmidrule(lr){10-12}
  & \textbf{Full} & \textbf{Author} & \textbf{Nest}
  & \textbf{Full} & \textbf{Author} & \textbf{Nest}
  & \textbf{Full} & \textbf{Author}
  & \textbf{Full} & \textbf{Author} & \textbf{Nest} \\
\midrule
\multicolumn{12}{c}{\textit{Previous / Fine-Tuned SOTA \citep{murzaku-rambow-2024-beleaf}}} \\
\midrule
GPT-3 (Fine-tuned) 
  & 65.8 & 76.0 & -- 
  & 65.8 & 76.0 & -- 
  & -- & -- 
  & -- & -- & -- \\
Flan-T5-XL         
  & 69.5 & 76.6 & -- 
  & 69.5 & 76.6 & -- 
  & -- & -- 
  & -- & -- & -- \\
\midrule
\multicolumn{12}{c}{\textit{Zero-Shot LLM Systems}} \\
\midrule
GPT-4o
  & 60.2 & 65.9 & 20.2
  & 68.7 & 73.2 & 22.9
  & \textcolor{red}{-0.8} & \textcolor{red}{-3.4}
  & {\textcolor{ForestGreen}{+8.5}} & {\textcolor{ForestGreen}{+7.3}} & {\textcolor{ForestGreen}{+2.7}} \\
o1\,\small\reasoning
  & 65.0 & \textbf{73.2} & 18.9
  & 70.3 & \textbf{78.9}\textsuperscript{$\dagger$} & 19.2
  & \textcolor{ForestGreen}{+0.8} & \textcolor{ForestGreen}{+2.3}
  & {\textcolor{ForestGreen}{+5.3}} & {\textcolor{ForestGreen}{+5.7}} & {\textcolor{ForestGreen}{+0.3}} \\
DeepSeek r1\,\small\reasoning\,\small\openmodel
  & \textbf{66.1} & 71.1 & \textbf{24.1}
  & \textbf{72.0}\textsuperscript{$\dagger$} & 77.6 & \textbf{25.3}\textsuperscript{$\dagger$}
  & \textcolor{ForestGreen}{+2.5} & \textcolor{ForestGreen}{+1.0}
  & {\textcolor{ForestGreen}{+5.9}} & {\textcolor{ForestGreen}{+6.5}} & {\textcolor{ForestGreen}{+1.2}} \\
o3-mini\,\small\reasoning
  & 62.4 & 70.9 & 15.6
  & 65.5 & 75.2 & 17.0
  & \textcolor{red}{-4.0} & \textcolor{red}{-1.4}
  & {\textcolor{ForestGreen}{+3.1}} & {\textcolor{ForestGreen}{+4.3}} & {\textcolor{ForestGreen}{+1.4}} \\
Claude 3.5
  & 63.2 & 69.7 & 19.7
  & 70.4 & 77.6 & 21.4
  & \textcolor{ForestGreen}{+0.9} & \textcolor{ForestGreen}{+1.0}
  & {\textcolor{ForestGreen}{+7.2}} & {\textcolor{ForestGreen}{+7.9}} & {\textcolor{ForestGreen}{+1.7}} \\
LLaMA 3.3\,\small\openmodel
  & 53.1 & 60.4 & 14.4
  & 58.8 & 66.0 & 19.9
  & \textcolor{red}{-10.7} & \textcolor{red}{-10.6}
  & {\textcolor{ForestGreen}{+5.7}} & {\textcolor{ForestGreen}{+5.6}} & {\textcolor{ForestGreen}{+5.5}} \\
DeepSeek-v3\,\small\openmodel
  & 56.3 & 61.4 & 17.1
  & 60.5 & 65.3 & 18.2
  & \textcolor{red}{-9.0} & \textcolor{red}{-11.3}
  & {\textcolor{ForestGreen}{+4.2}} & {\textcolor{ForestGreen}{+3.9}} & {\textcolor{ForestGreen}{+1.1}} \\
\bottomrule
\end{tabular}
\caption{\textbf{Unified} vs. \textbf{Hybrid} approaches with different LLMs. We report Micro F1 (Full), Author Micro F1 (Author), and Nested Micro F1 (Nest) scores (in \%). \textbf{\(\Delta\) Hyb.-SOTA} denotes the difference between the \textbf{Hybrid} result vs. the fine-tuned SOTA. The best scores are highlighted in \textbf{bold} and new state-of-the-art (SOTA) results are denoted by $\dagger$. \(\Delta\)~\textbf{Hyb.-Unif.} highlight the \textbf{Hybrid}-\textbf{Unified} difference for Full, Author, and Nest F1s. {\small\openmodel} indicates open models and {\small\reasoning} indicates reasoning models.
}
\label{tab:fb-results}
\end{table*}

\subsection{Evaluation: Metrics} 
\label{sec:methodology-evals}
We evaluate on three F1 metrics: \texttt{Full} where we perform an exact match evaluation on all generated (source, event, label) annotations; \texttt{Author} where we perform an evaluation on all generated annotations where the source is the author of the text; and \texttt{Nest} where we perform an exact match evaluation on all generated annotations where the source is a nested source.

 \subsection{Evaluation: FactBank Sources}
\label{methodology:normalization}
FactBank has specific conventions about annotating sources. Consider the example ``Trurit Inc. shares rose by 5\% today''. FactBank annotates on the token level, and the source is ``Inc.''. We do not wish to penalize LLMs for not knowing this conversion, and therefore propose a few-shot normalization technique for postprocessing.
We perform all source normalization experiments with GPT-4o \citep{gpt4o}.  
Exact prompts for our source normalization methods and a detailed ablation study are shown in Appendix~\ref{sec:appendix-source}.

Our task setup is as follows: Given a predicted source, we prompt GPT-4o to transform the predicted source
into a FactBank-compliant version.

\section{Results and Analysis}
\label{sec:results}

\subsection{FactBank} 
\noindent\textbf{Main Results} Our main results for FactBank are shown in Table~\ref{tab:fb-results}. We compare all our results to the previous fine-tuned SOTA from \citet{murzaku-rambow-2024-beleaf},
evaluating on exact match F1 (\texttt{Full}) and author exact match F1 (\texttt{Author}) as described in Section~\ref{sec:methodology-evals}. We add one more metric: nested exact match F1 (\texttt{Nest}), where we evaluate on nested sources only. 

Our \textbf{Unified} zero-shot results (column \textbf{Unified}) achieve competitive performance compared to fully fine-tuned models, with \textsc{r1} (66.1\% for \texttt{Full}) and o1 (73.2\% for \texttt{Author}). We outperform the fine-tuned GPT-3 model from \citet{murzaku-rambow-2024-beleaf} on \texttt{Full}, but do not outperform the Flan-T5-XL system. 

We achieve new SOTA
on FactBank with our \textbf{Hybrid} systems. Our \textsc{r1} \textbf{Hybrid} system achieves \texttt{Full} of 72.0\%, outperforming  the previous state of the art by 2.5\% (column $\Delta$ vs. SOTA). Similar to the \textbf{Unified} results, o1 excels in \texttt{Author}, achieving 78.9\% \texttt{Author} and outperforming the previous SOTA by 2.3\%. We also note that GPT-4o and Claude-3.5 also achieve competitive performance, with Claude-3.5 outperforming the previous SOTA on \texttt{Full} and \texttt{Author} by 0.9\% and 1.0\% respectively. We hypothesize that these models excel due to CoT prompting.

\noindent\textbf{Nest F1} We are the first to provide \texttt{Nest} F1 metrics on FactBank. Our top performing model is r1, which achieves a nested F1 of 25.3\%. For reasoning models o1, o3-mini, and r1, we notice that going from \textbf{Unified} to \textbf{Hybrid} does not increase \texttt{Nest} F1 dramatically (0.3\% for o1, 1.4\% for o3-mini, and 1.2\% for r1), showcasing the models' lack of capabilities
for nested belief predictions. We note that these results are low, and believe modelling of nested beliefs is essential future work and a challenging task for reasoning LLMs.

\noindent\textbf{Zero vs. Hybrid} We quantify the exact difference (in \%) between our \textbf{Unified} and \textbf{Hybrid} models in Table~\ref{tab:fb-results} (column \(\Delta\)~(Hyb.-Unif.)). 
We see improvements in every model, with the greatest improvements occuring in GPT-4o and Claude-3.5 for \texttt{Full} and \texttt{Author}.
On average, our \textbf{Hybrid} models outperform our \textbf{Unified} models by 5.7\% for \texttt{Full}, 5.9\% for \texttt{Author}, and 2.0\% for \texttt{Nest}. Our results emphasize the need for a specialized event tagger and hybrid approach, allowing the LLM to focus on linking sources and tagging belief labels. 
\begin{table}[t]
\centering
\begin{tabular}{lccc}
\toprule
\textbf{Model} & \textbf{Type} & \textbf{F1} \\
\midrule
DeBERTa & Fine-tuned & \textbf{89.0} \\
\midrule
\multirow{3}{*}{DeepSeek R1} & Zero-shot & 82.0 \\
                           & Few-shot  & 76.4 \\
\midrule
\multirow{3}{*}{GPT-4o}      & Zero-shot & 78.2 \\
                           & Few-shot  & 81.1 \\
\midrule
\multirow{3}{*}{Claude 3.5}      & Zero-shot & 83.3 \\
                           & Few-shot  & 81.8 \\
\bottomrule
\end{tabular}
\caption{Event detection performance (in \% F1) of various language models. The fine-tuned DeBERTa model outperforms all major LLMs in zero-shot and few-shot settings.}
\label{tab:event-detection}
\end{table}

\noindent\textbf{Event Detection} 
We investigate how LLMs perform on event tagging.  We show these results in Table~\ref{tab:event-detection}. We compare three LLMs (r1, GPT-4o, and Claude-3.5) to the fine-tuned DeBERTa event tagger used in the {\bf Hybrid} system.
For our LLMs, we try two configurations: zero-shot and few-shot (5 examples). We find that a fine-tuned DeBERTa outperforms all LLMs in all settings, emphasizing that event detection is 
still a difficult task. We leave further experimental details and prompts in Appendix~\ref{sec:appendix-event-tagger}.

\noindent\textbf{Error Analysis} We perform an error analysis on the top-performing model (\textsc{r1}, \textbf{Hybrid}) on nested beliefs (F1 of only 25.3\%). We categorize errors as follows: \textbf{(1) Source} mismatch, often labeling the author instead of the nested source or failing to classify pronoun sources such as “it” correctly (123 errors); \textbf{(2) FN} (false negatives on events), where context-dependent event nouns or verbs are missed (e.g., \emph{“acquisition,” “construction”}) (77 errors); \textbf{(3) FP} (false positives on events), over-predicting event nouns (73 errors); \textbf{(4) Label} errors, notably predicting \emph{True} or \emph{Probable} instead of \emph{Unknown} for future/uncommitted events (e.g., “Mary offered to \textbf{buy} an apple”, where the \textbf{buy} event should be \emph{Unknown}
) (53 errors). We note that the FN errors are consistent with findings from prior FactBank studies: \citet{murzaku-etal-2022-examining} also found similar errors. More detailed results and analysis of our error analysis are in Appendix~\ref{sec:appendix-ea}.

\begin{table}[t]
\setlength\tabcolsep{4pt}
\centering
\begin{tabular}{llc}
\toprule
\textbf{Model} & \textbf{Method} & \textbf{Bel.+Pol.} \\
\midrule
mT5 XXL      & Fine-tune & \textbf{64.4} \\
\midrule
DeepSeek r1  & Hybrid    & 63.6\textsuperscript{$\dagger$} \\
o3-mini      & Hybrid    & 62.6\textsuperscript{$\dagger$} \\
GPT-4o       & Hybrid    & 61.2\textsuperscript{$\dagger$} \\
\midrule
GPT-4o      & Unified      & 42.9 \\
o3-mini      & Unified      & 40.8 \\
DeepSeek r1  & Unified      & 38.6 \\
\bottomrule
\end{tabular}
\caption{Model performance on Belief+Polarity (Bel.+Pol.) F1. \citet{rovera-etal-2025-modafact} mT5-XXL baseline is shown in \textbf{bold} . Results on Bel.+Pol. metric within 5\% of the SOTA are marked with a \textsuperscript{$\dagger$}.}
\label{tab:moda-results}
\end{table}

\subsection{Multilingual Belief} 

The ModaFact Italian corpus \citep{rovera-etal-2025-modafact} annotates the author's belief, polarity, and modality towards events and temporal information.
We only use the belief and polarity annotations and combine these to tags similar to those of FactBank (and perform an exact match evaluation on Belief+Polarity F1).  This is different from how \citet{rovera-etal-2025-modafact} evaluate, but they kindly shared their raw results so that we could apply our evaluation.

\noindent\textbf{Results} We perform our ModaFact experiments with three cost-effective models that performed well for FactBank:
GPT-4o, o3-mini, and \textsc{r1}. Our results are shown in Table~\ref{tab:moda-results}. Unlike our FactBank results, we fall short of the fine-tuned SOTA for our \textbf{Hybrid} system (by 0.8\%).  Similar to our FactBank results, \textbf{Hybrid} strongly outperforms \textbf{Unified} in all settings.
Finally, we see that \textsc{r1}
and o3-mini (both reasoning models) come very close to the fine-tuned SOTA. GPT-4o also proves competitive, but falls short of the reasoning models by 2.4\%. We note that while we do not beat the SOTA, 
the LLMs we use are not explicitly trained for multilingual data (in contrast with mT5-XXL).  For example, \textsc{r1} is specifically optimized for English and Chinese data 
\citep{guo2025deepseek}. We hypothesize that future multilingual optimizations for these reasoning LLMs %
would in fact lead to a new SOTA for the ModaFact corpus.

\section{Conclusion}
\label{sec:conclusion}
We show that belief detection from text remains a challenging problem for LLMs.  This is particularly true for nested beliefs, which the author ascribes to other sources.  Our new SOTA system includes a distinct fine-tuned event detection component.

\section*{Limitations} 
While our model achieves a new state-of-the-art on the English only FactBank, our results, while still competitive, do not perform as well for the Italian ModaFact corpus. We acknowledge this as a shortcoming and aim to work towards broader multilingual generalization for this task.

We note that our LLM approach yields poor results on the nested F1 metric, indicating a large gap and potential for future improvement. We will explore improving these results in future work and believe this to be a gap for all major open, closed, and reasoning LLMs.

Finally, we note that our top performing LLM approach, while using the open DeepSeek r1 model, is reliant on API calls for the source normalization technique. We attempt to minimize costs by using GPT-4o, but note that we can (i) achieve better performance using a larger, reasoning model (more cost) or (ii) switch to an open model. We will explore both techniques.

\section*{Ethics Statement}
We note that our paper is foundational research and we are not tied to any direct applications. We do not foresee any potential risks with our work. We do not perform any annotations or human evaluation as we use the already existing FactBank dataset and ModaFact dataset.

\bibliography{anthology,custom}

\appendix

\section{Unified Experiments}
\label{sec:appendix-unified}
\paragraph{Details} For all \textbf{Unified} zero-shot experiments, we use a temperature of 0.0 where applicable (all models besides o1 and o3-mini). For o1 and o3-mini, we use the default reasoning setting (Medium). To prompt all other models (LLaMA-3.3, DeepSeek-v3, DeepSeek-r1, and Claude-3.5-Sonnet), we use the OpenRouter API \citep{openrouter}. The open models are ran at full precision (henceforth why we used the OpenRouter API and external providers). 

\paragraph{Prompt} Our zero-shot \textbf{Unified} prompt is shown in Figure~\ref{fig:zero-shot-prompt}.

\section{Hybrid Experiments}
\label{sec:appendix-hybrid}
\paragraph{Details} For all \textbf{Hybrid} zero-shot experiments, we use a temperature of 0.0 where applicable (all models besides o1 and o3-mini). For o1 and o3-mini, we use the default reasoning setting (Medium). To prompt all other models (LLaMA-3.3, DeepSeek-v3, DeepSeek-r1, and Claude-3.5-Sonnet), we use the OpenRouter API \citep{openrouter}. 

\paragraph{Prompt} Our \textbf{Hybrid} zero-shot prompt is shown in Figure~\ref{fig:hybrid-prompt}.

\section{LLM Experiment Details} For all our FactBank experiments, we report a single run, especially due to cost. We note that o1 experiments cost up to \$75 per run on the FactBank test set. To minimize randomness, we set the temperature to 0.0 where applicable (besides o1 and o3-mini). For o1 and o3-mini, we use the default reasoning setting (Medium). For our ModaFact experiments, we report the average over all five folds. Due to API costs and performing five-fold cross validation, we limit all ModaFact experiments to GPT-4o, o3-mini, and DeepSeek r1, which are the most cost effective models. 

\section{Dataset Details}
\label{sec:appendix-data}
\paragraph{FactBank} We use the author and source-and-target projection of FactBank from \citet{murzaku-rambow-2024-beleaf}, who follow the article split from \citet{murzaku-etal-2022-examining}. We use their provided code for data extraction and follow their exact article split. The release of the FactBank corpus that we use can be found at the Linguistic Data Consortium, catalog number LDC2009T23. The test set contains 280 sentences and 1,326 examples.

\paragraph{ModaFact} We use all five-folds of the test set of the ModaFact corpus from \citet{rovera-etal-2025-modafact}, which is publicly available. All results we report are averages over the five folds. To get the events from ModaFact for our \textbf{Hybrid} zero-shot experiments, we use the author's provided prediction files and inference script with mT5-XXL. 
\begin{table}[ht]
\centering
\begin{tabular}{lcc}
\hline
Fold   & Sentences & Examples \\
\hline
Fold 1 & 646       & 2098     \\
Fold 2 & 605       & 2097     \\
Fold 3 & 606       & 2096     \\
Fold 4 & 626       & 2094     \\
Fold 5 & 601       & 2090     \\
\hline
\end{tabular}
\caption{Dataset details for the ModaFact test set.}
\label{tab:modafact-appdx}
\end{table}

\section{Source Normalization}
\label{sec:appendix-source}
We propose two source normalization prompts: a few shot source normalization prompt and an oracle source normalization prompt. For these prompts, we use GPT-4o, with temperature 0.0. We prompt GPT-4o using the OpenAI API. Our exact few shot source normalization prompt is shown in Figure~\ref{fig:few-shot-norm}. Our exact oracle normalization prompt is shown in Figure~\ref{fig:source_normalization_prompt}.

\begin{table}[t]
\centering
\begin{tabular}{lcc}
\toprule
\textbf{Norm.} & \textbf{Full} & \textbf{Nest} \\
\midrule
None  & 68.9  & 17.5 \\
Few Shot  & 72.0\ & 25.3\ \\
Oracle   & \textbf{72.7}\ & \textbf{27.1} \\
\bottomrule
\end{tabular}
\caption{Performance of DeepSeek r1 (Hybrid) under three source normalization settings. “None” denotes no normalization, “Few Shot” applies few-shot normalization, and “Oracle” uses ground-truth for normalization. Bold values indicate the best results for each test set.}
\label{tab:deepseek-ablation}
\end{table}

\begin{table}[t]
\setlength\tabcolsep{4pt}
\centering
\small
\begin{tabular}{lr|lr}
\toprule
\textbf{Category} & \textbf{Count} & \textbf{Breakdown} & \textbf{Count} \\
\midrule
Source & 123 & Gold=AUTHOR & 50 \\
                  &     & Gold=``it''   & 13 \\
\midrule
FN   & 77  & Missed Noun     & 38 \\
                  &     & Missed Verb     & 30 \\
\midrule
Label  & 73  & Pred:CT+ $\rightarrow$ Gold:UU  & 28 \\
                  &     & Pred:PR+ $\rightarrow$ Gold:UU & 22 \\
\midrule
FP   & 53  & Predicted Noun & 33 \\
                  &     & Predicted Verb & 10 \\
\bottomrule
\end{tabular}
\caption{Error analysis for our \textbf{Hybrid} DeepSeek r1 system on nested predictions, showing counts of each error type relative to its category total count.}
\label{tab:error-analysis-percentages-only}
\end{table}
We perform an ablation analysis of our few-shot and oracle normalization techniques described in Section~\ref{methodology:normalization}. We showcase these results for our top performing system (DeepSeek r1, \textbf{Hybrid}) in Table~\ref{tab:deepseek-ablation}. Without any normalization, we achieve a Full F1 of 68.9\% and Nest F1 of 17.5\%. Our few shot normalization technique improves us 2.1\% for Full F1, and more notably by 7.8\% for Nest F1. Our oracle method, as expected (since we provide gold sources), performs even better than our few shot method, achieving a Full F1 of 72.7\% and 27.1\%. However, we choose to perform all experiments with our few shot normalization method instead of our oracle method to truly showcase LLMs capabilities for belief detection without any gold sources as input.

\section{Event Tagger}
\label{sec:appendix-event-tagger}
\paragraph{DeBERTa Tagger} We use DeBERTa-large for token classification, setting the number of labels to 2 (O vs. EVENT). We use the following hyperparameters: Epochs: 5; Batch Size: 16; Learning Rate: 1e-4; Max Sequence Length: 128. We do not perform any hyperparameter optimization or tuning. The model is trained using the HuggingFace Transformers library \citep{wolf-etal-2020-transformers}.

\paragraph{LLM Event Tagging} We perform event tagging on multiple LLMs. We set the temperature to 0.0. We use the OpenRouter API \citep{openrouter} for DeepSeek r1 and Claude 3.5, and the OpenAI API for GPT-4o. We do not perform experiments with o1 to avoid high costs. Our zero-shot event detection prompt is shown in Figure~\ref{fig:factbank_event_prompt_zs}. Our few-shot event detection prompt is shown in Figure~\ref{fig:factbank_few_shot_prompt_events}.

\section{Nested Error Analysis}
\label{sec:appendix-ea}

We expand our error analysis on the nested sources. 
Table~\ref{tab:error-analysis-percentages-only} shows our error counts and error types.

We specifically analyze the errors for nested beliefs, which is where all LLMs fail on (our top performing model achieving F1 of only 25.3\%). We showcase the error category, and then the top two error types by count. We use the following labels: {\bf Source} indicates a source mismatch error; {\bf FN} indicates a false negative, where the LLM did not generated a certain event type; {\bf Label} indicates a label error where the LLM had the source and event correct, but inorrectly labeled the event. {\bf FP} indicates a false positive, where the LLM overpredicted (that is, it generated an event that was not actually an event). 

Our most notable error is {\bf Source}, where 123 errors are made. The most common error is the model predicts the author is the source instead of the correct nested source. Another notable error is where the model does not classify the source as it, but rather predicts the name of the entity. Next, we see a repeating of similar errors that \citet{murzaku-etal-2022-examining} discovered. Specifically, event nouns can be hard to determine (e.g. nouns like ``concerns'', ``acquisition'', ``construction''). Our {\bf FN}  and {\bf FP} errors showcase that LLMs simultaneously over predict event nouns, while also missing both event nouns and verbs. Finally, we notice two notable label flips for our {\bf Label} error category: the LLM predicts CT+ (the event happened/is true) when the gold label is UU (unknown), and PR+ (possibly true) when the gold is UU. This is due to FactBank's definitions of nested sources and future events: when a reporting of a future event happening (e.g. ``Mary said it will happen''), the factuality of the event according to the source is UU (the source is not committing to the event; rather, the author is commiting to it). 

Our analysis emphasizes that despite our source normalization method and use of strong reasoning LLMs, there is much room for improvement. Our error analysis findings are further supported with similar errors that have been reported in previous works on FactBank \citep{murzaku-etal-2022-examining}.

\section{Code Release} We will release all of our code. We will provide the full pipelines, datasets, and model checkpoints where applicable. 

\definecolor{Lavender}{HTML}{E6E6FA}
\definecolor{Periwinkle}{HTML}{CCCCFF}
\definecolor{SeaGreen}{RGB}{46, 139, 87}      %
\definecolor{SkyBlue}{RGB}{135, 206, 235}     %
\definecolor{Orange}{HTML}{fdae61}
\definecolor{Brown}{HTML}{dfc27d}

\begin{figure*}[!ht]
\centering
\small
\caption{\textbf{Instruction for our zero-shot Unified Belief Annotation.} The instruction for FactBank-style event factuality annotation consists of three parts: a brief \hlc[SeaGreen!10]{task description}, detailed \hlc[Periwinkle!30]{step-by-step instructions}, and the \hlc[SkyBlue!20]{formatting structure}. Our CoT instructions are shown in the end of the prompt (Step-by-Step Output).}
\begin{tcolorbox}[
    width=\textwidth,
    colback=white,
    colframe=black,
    arc=4mm,
    boxrule=0.5pt,
    left=2mm,
    right=2mm,
    top=2mm,
    bottom=2mm,
    fonttitle=\bfseries,
    ]

\begin{tcolorbox}[
    colback=SeaGreen!8,
    boxrule=0pt,
    colframe=white,
    left=0pt,
    right=0pt,
    top=0pt,
    bottom=0pt,
    ]

\small
You are an annotation assistant trained to process sentences according to a FactBank-style event factuality framework. Given a sentence (or short text), your task is to analyze and annotate events by:

\begin{itemize}[noitemsep, leftmargin=15pt, topsep=0pt]
    \item \textbf{Event Identification}: Finding and listing all event-denoting predicates (verbs, event nouns, state-denoting adjectives)
    \item \textbf{Source Analysis}: Identifying who is expressing or committing to each event
    \item \textbf{Factuality Assessment}: Determining how certain each source is about the events
    \item \textbf{Nested Attribution}: Managing multiple layers of reporting and belief
    \item \textbf{Special Cases}: Handling future events, negation, modality, and hedging
\end{itemize}
\end{tcolorbox}

\begin{tcolorbox}[
    colback=Periwinkle!20,
    boxrule=0pt,
    colframe=white,
    left=0pt,
    right=0pt,
    top=0pt,
    bottom=0pt,
    ]

\small
Follow these steps precisely for annotation:\\

\textbf{STEP 1: Event Identification}
\begin{itemize}[noitemsep, leftmargin=15pt, topsep=0pt]
    \item Find all event-denoting predicates in the text
    \item Each predicate must be a single token
    \item Include verbs, event nouns, and state-denoting adjectives
\end{itemize}

\textbf{STEP 2: Source Identification}
\begin{itemize}[noitemsep, leftmargin=15pt, topsep=0pt]
    \item Start with "AUTHOR" as the root source (text narrator)
    \item Identify source-introducing predicates (SIPs) like "said," "believed," "reported"
    \item For new sources (e.g., "Apple officials"), normalize to single-token labels (e.g., "officials")
    \item Format as "AUTHOR\_<ShortLabel>" (e.g., "AUTHOR\_officials")
    \item For nested sources, add additional levels with underscores (e.g., "AUTHOR\_officials\_spokesperson")
    \item Handle negated sources (e.g., "did not say") at the higher level
\end{itemize}

\textbf{STEP 3: Factuality Labeling}
Assign one of these labels for each event-source pair:
\begin{itemize}[noitemsep, leftmargin=15pt, topsep=0pt]
    \item \textbf{true}: Certainly factual (e.g., "confirmed," "knew")
    \item \textbf{false}: Certainly counterfactual (e.g., "denied," "did not happen")
    \item \textbf{ptrue}: Probably true (e.g., "might," "could," "likely")
    \item \textbf{pfalse}: Probably false (e.g., "doubted")
    \item \textbf{unknown}: Non-committal or unspecified stance
\end{itemize}

\textbf{Special Cases Guidelines}
\begin{itemize}[noitemsep, leftmargin=15pt, topsep=0pt]
    \item Future/prospective events: Label as unknown unless probability indicated (then ptrue)
    \item Negative statements: Use false for explicit denials
    \item Modality/hedging: Use ptrue for "might," "could," "suspected"
    \item Uncommitted author: Use unknown for purely reported events
\end{itemize}

\end{tcolorbox}

\begin{tcolorbox}[
    colback=SkyBlue!10,
    boxrule=0pt,
    colframe=white,
    left=0pt,
    right=0pt,
    top=0pt,
    bottom=0pt,
    ]

\small
Your annotation should be formatted as a JSON-style list of dictionaries:

\begin{verbatim}
[
  {
    "source": "<source_label>",  // e.g., "AUTHOR" or "AUTHOR_<source>"
    "event": "<event_token>",    // exact predicate from text
    "label": "<factuality_value>" // true/false/ptrue/pfalse/unknown
  },
  ...
]
\end{verbatim}

Step-by-Step Output Process:
\begin{itemize}[noitemsep, leftmargin=15pt, topsep=0pt]
    \item Walk through each event in the sentence
    \item Identify and explain all sources and their nesting
    \item Justify each factuality label from each source's viewpoint
    \item Produce the final JSON-style output
\end{itemize}

\end{tcolorbox}

\end{tcolorbox}
\label{fig:zero-shot-prompt}
\end{figure*}

\definecolor{Lavender}{HTML}{E6E6FA}
\definecolor{Periwinkle}{HTML}{CCCCFF}
\definecolor{SeaGreen}{RGB}{46, 139, 87}      %
\definecolor{SkyBlue}{RGB}{135, 206, 235}     %
\definecolor{Orange}{HTML}{fdae61}
\definecolor{Brown}{HTML}{dfc27d}

\begin{figure*}[!ht]
\centering
\small
\caption{\textbf{Instruction for our Hybrid Belief Annotation.} The instruction for FactBank-style event factuality annotation consists of three parts: a brief \hlc[SeaGreen!10]{task description}, detailed \hlc[Periwinkle!30]{step-by-step instructions}, and the \hlc[SkyBlue!20]{formatting structure}. Our CoT instructions are shown in the end of the prompt (Step-by-Step Output).}
\begin{tcolorbox}[
    width=\textwidth,
    colback=white,
    colframe=black,
    arc=4mm,
    boxrule=0.5pt,
    left=2mm,
    right=2mm,
    top=2mm,
    bottom=2mm,
    fonttitle=\bfseries,
    ]

\begin{tcolorbox}[
    colback=SeaGreen!8,
    boxrule=0pt,
    colframe=white,
    left=0pt,
    right=0pt,
    top=0pt,
    bottom=0pt,
    ]

\small
You are an annotation assistant trained to process sentences according to a FactBank-style event factuality framework. Given a sentence (or short text) and a list of event predicates marked in that sentence, your task is to analyze and annotate events by:

\begin{itemize}[noitemsep, leftmargin=15pt, topsep=0pt]
    \item \textbf{Source Analysis}: Identifying who is expressing or committing to each event
    \item \textbf{Factuality Assessment}: Determining how certain each source is about the events
    \item \textbf{Nested Attribution}: Managing multiple layers of reporting and belief
\end{itemize}
\end{tcolorbox}

\begin{tcolorbox}[
    colback=Periwinkle!20,
    boxrule=0pt,
    colframe=white,
    left=0pt,
    right=0pt,
    top=0pt,
    bottom=0pt,
    ]

\small
Follow these steps precisely for annotation:\\

\textbf{STEP 1: Source Identification}
\begin{itemize}[noitemsep, leftmargin=15pt, topsep=0pt]
    \item Start with "AUTHOR" as the root source (text narrator)
    \item Identify source-introducing predicates (SIPs) like "said," "believed," "reported," "estimated," "argued"
    \item For new sources (e.g., "Apple officials"), normalize to single-token labels (e.g., "officials")
    \item Format as "AUTHOR\_<ShortLabel>" (e.g., "AUTHOR\_officials")
    \item For nested sources, add additional levels with underscores (e.g., "AUTHOR\_officials\_spokesperson")
    \item Handle negated sources (e.g., "did not say") at the higher level
\end{itemize}

\textbf{STEP 2: Factuality Labeling}
Assign one of these labels for each event-source pair:
\begin{itemize}[noitemsep, leftmargin=15pt, topsep=0pt]
    \item \textbf{true}: Certainly factual (e.g., "confirmed," "knew")
    \item \textbf{false}: Certainly counterfactual (e.g., "denied," "did not happen")
    \item \textbf{ptrue}: Probably true (e.g., "might," "could," "likely")
    \item \textbf{pfalse}: Probably false (e.g., "doubted")
    \item \textbf{unknown}: Non-committal or unspecified stance
\end{itemize}

\textbf{Special Cases Guidelines}
\begin{itemize}[noitemsep, leftmargin=15pt, topsep=0pt]
    \item Future/prospective events: Label as unknown unless probability indicated (then ptrue)
    \item Negative statements: Use false for explicit denials
    \item Modality/hedging: Use ptrue for "might," "could," "suspected"
    \item Uncommitted author: Use unknown for purely reported events
\end{itemize}

\end{tcolorbox}

\begin{tcolorbox}[
    colback=SkyBlue!10,
    boxrule=0pt,
    colframe=white,
    left=0pt,
    right=0pt,
    top=0pt,
    bottom=0pt,
    ]

\small
Your annotation should be formatted as a JSON-style list of dictionaries:

\begin{verbatim}
[
  {
    "source": "<source_label>",  // e.g., "AUTHOR" or "AUTHOR_<source>"
    "event": "<event_token>",    // exact predicate from text
    "label": "<factuality_value>" // true/false/ptrue/pfalse/unknown
  },
  ...
]
\end{verbatim}

Step-by-Step Output Process:
\begin{itemize}[noitemsep, leftmargin=15pt, topsep=0pt]
    \item Walk through each event in the sentence
    \item Identify and explain all sources and their nesting
    \item Justify each factuality label from each source's viewpoint
    \item Produce the final JSON-style output
\end{itemize}

\end{tcolorbox}
\end{tcolorbox}
\label{fig:hybrid-prompt}
\end{figure*}

\definecolor{Lavender}{HTML}{E6E6FA}
\definecolor{Periwinkle}{HTML}{CCCCFF}
\definecolor{SeaGreen}{RGB}{46, 139, 87}
\definecolor{SkyBlue}{RGB}{135, 206, 235}
\definecolor{Orange}{HTML}{fdae61}
\definecolor{Brown}{HTML}{dfc27d}

\begin{figure*}[!ht]
\centering
\small
\caption{\textbf{Oracle Source Normalization Prompt}}
\begin{tcolorbox}[
    width=\textwidth,
    colback=white,
    colframe=black,
    arc=4mm,
    boxrule=0.5pt,
    left=2mm,
    right=2mm,
    top=2mm,
    bottom=2mm,
    fonttitle=\bfseries,
    ]
\begin{tcolorbox}[
    colback=SeaGreen!8,
    boxrule=0pt,
    colframe=white,
    left=0pt,
    right=0pt,
    top=0pt,
    bottom=0pt,
    ]
    
\small
You are determining if two source names refer to the same entity.
Consider company abbreviations, common variations, and parent/subsidiary relationships. Also consider the context of the sentence and entity coreference.
\bigskip

Answer only YES if these definitely refer to the same entity, NO if they are different or if you're unsure.
Include a brief explanation of your reasoning.
\begin{tcolorbox}[
    colback=SkyBlue!10,
    boxrule=0pt,
    colframe=white,
    left=0pt,
    right=0pt,
    top=0pt,
    bottom=0pt,
    ]
\begin{verbatim}
Sentence: {Sentence}
Predicted Source: {Predicted Source}
Gold Source: {Gold Source}
\end{verbatim}
\end{tcolorbox}
\end{tcolorbox}
\end{tcolorbox}
\label{fig:source_normalization_prompt}
\end{figure*}

\definecolor{Lavender}{HTML}{E6E6FA}
\definecolor{Periwinkle}{HTML}{CCCCFF}
\definecolor{SeaGreen}{RGB}{46, 139, 87}
\definecolor{SkyBlue}{RGB}{135, 206, 235}
\definecolor{Orange}{HTML}{fdae61}
\definecolor{Brown}{HTML}{dfc27d}

\begin{figure*}[!ht]
\centering
\small
\caption{\textbf{Few Shot Source Normalization Prompt}}
\begin{tcolorbox}[
    width=\textwidth,
    colback=white,
    colframe=black,
    arc=4mm,
    boxrule=0.5pt,
    left=2mm,
    right=2mm,
    top=2mm,
    bottom=2mm,
    fonttitle=\bfseries,
    ]
    
\begin{tcolorbox}[
    colback=SeaGreen!8,
    boxrule=0pt,
    colframe=white,
    left=0pt,
    right=0pt,
    top=0pt,
    bottom=0pt,
    ]
\small
You are a FactBank-style source normalization assistant.

Your task: \textbf{Identify and normalize the subject (speaker/thinker/etc.) of each source-introducing predicate (SIP)} in a sentence. The normalized form must be a short, single-token label following ``AUTHOR\_''. If nested sources appear (i.e., one speaker quotes another speaker), nest them by appending an underscore plus the new label.
\end{tcolorbox}

\begin{tcolorbox}[
    colback=Periwinkle!20,
    boxrule=0pt,
    colframe=white,
    left=0pt,
    right=0pt,
    top=0pt,
    bottom=0pt,
    ]
\small
Use these \textbf{rules and guidelines}:

1. \textbf{Source-Introducing Predicates (SIPs)}:
   \begin{itemize}[noitemsep, leftmargin=15pt]
      \item Common SIP verbs: ``said,'' ``reported,'' ``believed,'' ``estimated,'' ``argued,'' ``announced,'' ``denied,'' ``claimed,'' etc.
      \item If an entity is repeated (the same subject for multiple SIPs), reuse the same label.
   \end{itemize}

2. \textbf{Normalization}:
   \begin{itemize}[noitemsep, leftmargin=15pt]
      \item Reduce corporate entities to ``Corp.'' or ``Inc.'' instead of the full name. E.g.:
         \begin{itemize}[noitemsep, leftmargin=15pt]
            \item ``Marathon Widget Corp.'' $\rightarrow$ \textbf{AUTHOR\_Corp.}
            \item ``Skyline Media Inc.'' $\rightarrow$ \textbf{AUTHOR\_Inc.}
         \end{itemize}
      \item If it's just ``the company,'' consider normalizing to \textbf{AUTHOR\_company} \textit{only} if no more specific corporate form (like ``Corp.'') is available.
      \item For people:
         \begin{itemize}[noitemsep, leftmargin=15pt]
            \item ``He,'' ``she,'' ``they'' $\rightarrow$ \textbf{AUTHOR\_he}, \textbf{AUTHOR\_she}, \textbf{AUTHOR\_they}
            \item ``Mr. Alvarez,'' ``Ms. Hurt,'' or ``Dr. Kim'' $\rightarrow$ \textbf{AUTHOR\_Alvarez}, \textbf{AUTHOR\_Hurt}, \textbf{AUTHOR\_Kim}
            \item If the sentence says ``I stated...'' $\rightarrow$ \textbf{AUTHOR\_I}
         \end{itemize}
      \item For large institutions:
         \begin{itemize}[noitemsep, leftmargin=15pt]
            \item ``Ministry of Defense'' $\rightarrow$ \textbf{AUTHOR\_ministry}
            \item ``Police Department'' $\rightarrow$ \textbf{AUTHOR\_police}
            \item ``officials'' $\rightarrow$ \textbf{AUTHOR\_officials}
            \item ``board'' $\rightarrow$ \textbf{AUTHOR\_board}
         \end{itemize}
      \item If you have a nested quote, e.g., ``AUTHOR\_officials\_spokesperson'' if the spokesperson is quoting officials.
   \end{itemize}

3. \textbf{Polarity}:
   \begin{itemize}[noitemsep, leftmargin=15pt]
      \item Even if the SIP is negated, you still label that source. (e.g., ``he denied...'' is valid.)
   \end{itemize}

4. \textbf{Output}:
   \begin{itemize}[noitemsep, leftmargin=15pt]
      \item Output \textbf{only} the normalized label(s). If no new source is introduced, or if you're uncertain, you can leave the text unchanged or indicate ``No SIP found.''
   \end{itemize}
\end{tcolorbox}

\bigskip

\textbf{Few-Shot Examples} \textit{(truncated for space; we use 10 few shot examples)}

\begin{enumerate}[noitemsep]
   \item \textbf{Sentence}: Alpha Widget Corp. said it is launching a new product line.\\
         \textbf{Predicted}: AUTHOR\_Alpha\_Widget\_Corp.\\
         \textbf{Corrected}: AUTHOR\_Corp.
   \item \textbf{Sentence}: ``I believe the results speak for themselves,'' he announced.\\
         \textbf{Predicted}: AUTHOR\_he\\
         \textbf{Corrected}: AUTHOR\_I\\
         (\textit{Because ``I'' is the direct speaker—if the text clearly attributes the quote to the first person.})
   \item \textbf{Sentence}: In its quarterly filing, LRS Acquisition stated it expects higher revenue.\\
         \textbf{Predicted}: AUTHOR\_LRS\\
         \textbf{Corrected}: AUTHOR\_Acquisition
   \item \textbf{Sentence}: A portfolio unit of Greenbank Corp. reported continued growth this year.\\
         \textbf{Predicted}: AUTHOR\_portfolio unit\\
         \textbf{Corrected}: AUTHOR\_unit
   \item \textbf{Sentence}: The foreign minister declared that cooperation would improve global stability.\\
         \textbf{Predicted}: AUTHOR\_foreign minister\\
         \textbf{Corrected}: AUTHOR\_minister
\end{enumerate}
\begin{tcolorbox}[
    colback=SkyBlue!10,
    boxrule=0pt,
    colframe=white,
    left=0pt,
    right=0pt,
    top=0pt,
    bottom=0pt,
    ]
\small
\textbf{Return}:
\begin{itemize}[noitemsep, leftmargin=15pt]
   \item Return the final normalized label(s) if a new source arises from the SIP.
   \item If none or unclear, output ``No SIP found'' or leave the text as is.
\end{itemize}
\end{tcolorbox}

\end{tcolorbox}
\label{fig:few-shot-norm}
\end{figure*}

\definecolor{Lavender}{HTML}{E6E6FA}
\definecolor{Periwinkle}{HTML}{CCCCFF}
\definecolor{SeaGreen}{RGB}{46, 139, 87}
\definecolor{SkyBlue}{RGB}{135, 206, 235}
\definecolor{Orange}{HTML}{fdae61}
\definecolor{Brown}{HTML}{dfc27d}

\begin{figure*}[!ht]
\centering
\small
\caption{\textbf{FactBank Single-Token Event Identification Prompt}}
\begin{tcolorbox}[
    width=\textwidth,
    colback=white,
    colframe=black,
    arc=4mm,
    boxrule=0.5pt,
    left=2mm,
    right=2mm,
    top=2mm,
    bottom=2mm,
    fonttitle=\bfseries,
    ]
    
\begin{tcolorbox}[
    colback=SeaGreen!8,
    boxrule=0pt,
    colframe=white,
    left=0pt,
    right=0pt,
    top=0pt,
    bottom=0pt,
    ]
\small
You are an expert at identifying single-token events in text following FactBank guidelines.

Find ALL single-token predicates that:
\end{tcolorbox}
\begin{tcolorbox}[
    colback=Periwinkle!20,
    boxrule=0pt,
    colframe=white,
    left=0pt,
    right=0pt,
    top=0pt,
    bottom=0pt,
    ]
\small
\textbf{Criteria:}
\begin{enumerate}[noitemsep]
    \item \textbf{Are ONLY ONE of:}
    \begin{itemize}[noitemsep, leftmargin=15pt]
        \item Reporting verbs (communication)
        \item Cognitive verbs (mental states)
        \item Action verbs (physical/abstract actions)
        \item Event nouns (occurrences/happenings)
        \item State adjectives (temporary states)
    \end{itemize}
    \item \textbf{Must represent:}
    \begin{itemize}[noitemsep, leftmargin=15pt]
        \item Something that happened/happens/will happen
        \item Something that can be assessed as true or false
        \item Something with a temporal dimension
    \end{itemize}
    \item \textbf{Critical Distinction for Nouns/Nominalizations:}
    \begin{itemize}[noitemsep, leftmargin=15pt]
        \item \textbf{INCLUDE} only nouns that refer to specific instances of events with:
        \begin{itemize}[noitemsep, leftmargin=20pt]
            \item Concrete temporal bounds
            \item Specific participants
            \item Ability to be assessed as having occurred or not
        \end{itemize}
        \item \textbf{DO NOT include} nouns that refer to:
        \begin{itemize}[noitemsep, leftmargin=20pt]
            \item General concepts or types of events
            \item Abstract categories
            \item Topics or subjects of discussion
            \item Generic processes
            \item Institutional practices
        \end{itemize}
    \end{itemize}
\end{enumerate}
\textbf{Key rules:}
\begin{itemize}[noitemsep, leftmargin=15pt]
    \item Extract SINGLE tokens only
    \item Include all verbs from source-introducing predicates 
    \item Include nested events 
    \item Include events under modals or negation
    \item Include events in complement clauses
\end{itemize}
\textbf{Do NOT include:}
\begin{itemize}[noitemsep, leftmargin=15pt]
    \item Multi-word phrases
    \item Generic nouns 
    \item Auxiliary verbs
    \item Articles, prepositions, or conjunctions
    \item References to event types without specific instances
\end{itemize}
\end{tcolorbox}
\begin{tcolorbox}[
    colback=SkyBlue!10,
    boxrule=0pt,
    colframe=white,
    left=0pt,
    right=0pt,
    top=0pt,
    bottom=0pt,
    ]
\small
\textbf{Output Format:}

Your output is a JSON-style list of dictionaries. Each dictionary has:
\begin{itemize}[noitemsep, leftmargin=15pt]
    \item \texttt{"event"}: The exact event token or predicate from the sentence.
\end{itemize}
\textbf{Output Example:}
\begin{verbatim}
[
  {"event": "EVENT1"},
  {"event": "EVENT2"},
  {"event": "EVENT3"}
]
\end{verbatim}
\end{tcolorbox}

\end{tcolorbox}
\label{fig:factbank_event_prompt_zs}
\end{figure*}

\definecolor{Lavender}{HTML}{E6E6FA}
\definecolor{Periwinkle}{HTML}{CCCCFF}
\definecolor{SeaGreen}{RGB}{46, 139, 87}
\definecolor{SkyBlue}{RGB}{135, 206, 235}
\definecolor{Orange}{HTML}{fdae61}
\definecolor{Brown}{HTML}{dfc27d}

\begin{figure*}[!ht]
\centering
\small
\caption{\textbf{FactBank Few-Shot Single-Token Event Identification Prompt.}}
\begin{tcolorbox}[
    width=\textwidth,
    colback=white,
    colframe=black,
    arc=4mm,
    boxrule=0.5pt,
    left=2mm,
    right=2mm,
    top=2mm,
    bottom=2mm,
    fonttitle=\bfseries,
    ]
    
\begin{tcolorbox}[
    colback=SeaGreen!8,
    boxrule=0pt,
    colframe=white,
    left=0pt,
    right=0pt,
    top=0pt,
    bottom=0pt,
    ]
\small
You are an expert at identifying single-token events in text following FactBank guidelines.

Find ALL single-token predicates that:
\end{tcolorbox}
\begin{tcolorbox}[
    colback=Periwinkle!20,
    boxrule=0pt,
    colframe=white,
    left=0pt,
    right=0pt,
    top=0pt,
    bottom=0pt,
    ]
\small
1. \textbf{Are ONLY ONE of:}
\begin{itemize}[noitemsep, leftmargin=15pt]
  \item Reporting verbs (communication)
  \item Cognitive verbs (mental states)
  \item Action verbs (physical/abstract actions)
  \item Event nouns (occurrences/happenings)
  \item State adjectives (temporary states)
\end{itemize}

2. \textbf{Must represent:}
\begin{itemize}[noitemsep, leftmargin=15pt]
  \item Something that happened/happens/will happen
  \item Something that can be assessed as true or false
  \item Something with a temporal dimension
\end{itemize}

3. \textbf{Critical Distinction for Nouns/Nominalizations:}
\begin{itemize}[noitemsep, leftmargin=15pt]
  \item \textbf{INCLUDE} only nouns that refer to specific instances of events with:
  \begin{itemize}[noitemsep, leftmargin=20pt]
    \item Concrete temporal bounds
    \item Specific participants
    \item Ability to be assessed as having occurred or not
  \end{itemize}
  \item \textbf{DO NOT include} nouns that refer to:
  \begin{itemize}[noitemsep, leftmargin=20pt]
    \item General concepts or types of events
    \item Abstract categories
    \item Topics or subjects of discussion
    \item Generic processes
    \item Institutional practices
  \end{itemize}
\end{itemize}

\textbf{Key rules:}
\begin{itemize}[noitemsep, leftmargin=15pt]
  \item Extract SINGLE tokens only
  \item Include all verbs from source-introducing predicates 
  \item Include nested events 
  \item Include events under modals or negation
  \item Include events in complement clauses
\end{itemize}
\textbf{Do NOT include:}
\begin{itemize}[noitemsep, leftmargin=15pt]
  \item Multi-word phrases
  \item Generic nouns 
  \item Auxiliary verbs
  \item Articles, prepositions, or conjunctions
  \item References to event types without specific instances
\end{itemize}
\end{tcolorbox}
\begin{tcolorbox}[
    colback=SkyBlue!10,
    boxrule=0pt,
    colframe=white,
    left=0pt,
    right=0pt,
    top=0pt,
    bottom=0pt,
    ]
\small
\textbf{Output Format:}

Your output is a JSON-style list of dictionaries. Each dictionary has:
\begin{itemize}[noitemsep, leftmargin=15pt]
  \item \texttt{"event"}: The exact event token or predicate from the sentence.
\end{itemize}

\textbf{Examples: (we truncate the examples omit 3 examples here for brevity)}

1. \textbf{Sentence:} In composite trading Friday on the New York Stock Exchange, BellSouth shares fell 87.5 cents.\\
\textbf{Output:}
\begin{verbatim}
[
  {"event": "trading"},
  {"event": "fell"},
]
\end{verbatim}

2. \textbf{Sentence:} Many local residents denounced the bigotry.\\
\textbf{Output:}
\begin{verbatim}
[
   {"event": "denounced"},
   {"event": "bigotry"}
]
\end{verbatim}

\end{tcolorbox}
\end{tcolorbox}
\label{fig:factbank_few_shot_prompt_events}
\end{figure*}

\end{document}